\documentclass[11pt]{article}
\usepackage[hyperref]{acl}
\usepackage{times}
\usepackage{latexsym}
\usepackage[T1]{fontenc}
\usepackage[utf8]{inputenc}
\usepackage{microtype}
\usepackage{booktabs}
\usepackage{multirow}
\usepackage{graphicx}
\usepackage{pgfplots}
\usepackage{xcolor}
\pgfplotsset{compat=1.18}

\definecolor{engfirst}{RGB}{31,119,180}
\definecolor{bilingual}{RGB}{255,127,14}
\definecolor{multilingual}{RGB}{44,160,44}

\title{Left Behind: Cross-Lingual Transfer as a Bridge for Low-Resource Languages in Large Language Models}

\author{Abdul-Salem Beibitkhan \\
  North American University \\
  \texttt{abyeibitkhan@na.edu}}

\begin{document}
\maketitle

\begin{abstract}
We investigate how large language models perform on low-resource languages by benchmarking eight LLMs across five experimental conditions in English, Kazakh, and Mongolian. Using 50 hand-crafted questions spanning factual, reasoning, technical, and culturally grounded categories, we evaluate 2,000 responses on accuracy, fluency, and completeness. We find a consistent performance gap of 13.8--16.7 percentage points between English and low-resource language conditions, with models maintaining surface-level fluency while producing significantly less accurate content. Cross-lingual transfer---prompting models to reason in English before translating back---yields selective gains for bilingual architectures (+2.2pp to +4.3pp) but provides no benefit to English-dominant models. Our results demonstrate that current LLMs systematically underserve low-resource language communities, and that effective mitigation strategies are architecture-dependent rather than universal.
\end{abstract}

\section{Introduction}

Large language models are increasingly deployed as general-purpose tools across linguistic boundaries, yet their capabilities remain profoundly unequal across languages. While English and other high-resource languages benefit from vast training corpora and extensive evaluation benchmarks, low-resource languages (LRLs), spoken by hundreds of millions, receive a fraction of this attention \citep{joshi2020state, nllb2022no}. For users who primarily communicate in languages such as Kazakh ({\raise.17ex\hbox{$\scriptstyle\sim$}}19M speakers) or Mongolian ({\raise.17ex\hbox{$\scriptstyle\sim$}}6M speakers), this asymmetry translates into degraded accuracy, unreliable fluency, and a fundamentally inferior experience compared to their English-speaking counterparts.

Existing multilingual evaluations have largely focused on well-studied language families \citep{ahuja2023mega, bandarkar2024belebele} or relied on machine-translated benchmarks that fail to capture culturally grounded knowledge. Turkic and Mongolic languages remain particularly underexplored despite their typological diversity---agglutinative morphology, Cyrillic and Latin script variation, and rich cultural traditions that resist translation from English-centric training data---with recent pilot studies only beginning to address this gap \citep{maxutov2024llms, zhang2024mmeval}. Meanwhile, cross-lingual transfer (CLT)---prompting models to translate a question into English, reason in English, and translate the answer back---has been proposed as a practical workaround for LRL limitations \citep{shi2023language}, but its effectiveness across model architectures remains largely untested.

We present a controlled benchmark of 2,000 responses spanning eight LLMs, five experimental conditions, and 50 hand-crafted questions with human-authored parallel versions in English, Kazakh, and Mongolian. We evaluate models across three architecture tiers---English-first, Bilingual, and Multilingual---and compare direct prompting against an explicit CLT pipeline. Our findings reveal a consistent performance gap of 13.8--16.7 percentage points between English and LRL conditions, a selective CLT benefit that favors bilingual models (+2.2pp to +4.3pp) over English-first architectures, and a striking failure of a purpose-built multilingual model that produces Kyrgyz instead of Kazakh. These results suggest that the current generation of LLMs systematically underserves low-resource language communities, and that cross-lingual transfer offers a partial but architecturally dependent remedy. \footnote{All data, code, and benchmark questions are available at \url{https://github.com/abdulsal3m/left-behind-clt}.}

\section{Methodology}

\paragraph{Benchmark.}
We construct a 50-question benchmark spanning five categories of ten questions each: neutral factual, neutral reasoning, neutral technical, Kazakh-aligned cultural, and Mongolian-aligned cultural. The neutral categories test language-agnostic capabilities (knowledge retrieval, logical inference, domain vocabulary), while the cultural categories probe language-specific knowledge that may be underrepresented in English-centric training corpora. Each question exists in three human-translated parallel versions---English, Kazakh, and Mongolian---with a reference answer sheet permitting multiple acceptable responses where ambiguity exists.

\paragraph{Models.}
We evaluate eight generative large language models grouped by training data profile. \textit{English-first} models (Claude Opus 4.5, GPT-5.2, Gemini 3 Pro, Grok 4.1, Llama 4 Maverick) are predominantly trained on English corpora with multilingual capabilities as a secondary outcome. \textit{Bilingual} models (Qwen3, DeepSeek V3.2) incorporate substantial Chinese and multilingual data alongside English, hypothesized to transfer better to non-Latin scripts. \textit{Multilingual} models (Aya Expanse) are explicitly trained for broad language coverage spanning 100+ languages, including Turkic family languages \citep{ustun2024aya}.

\paragraph{Conditions.}
We define five experimental conditions. C1 (English baseline) presents questions and elicits responses in English. C2 and C4 (direct) present questions in Kazakh and Mongolian respectively, instructing models to respond in the target language. C3 and C5 (cross-lingual transfer) employ a three-step pipeline prompted entirely in the target language: (1) translate the question to English, (2) answer in English, (3) translate the answer back to the target language. Each of the 2,000 model--question pairs (8 models $\times$ 5 conditions $\times$ 50 questions) uses an independent conversation with no prior context.

\paragraph{Evaluation.}
Responses are scored on three dimensions---Accuracy (0--2), Fluency (0--2), and Completeness (0--2)---for a maximum of 6 points per response. Grading is performed semi-automatically using Claude Sonnet 4.5, following the LLM-as-judge paradigm \citep{zheng2023judging}, to streamline grading for black-and-white questions with the complete reference answer sheet provided as context. Fluency scoring enforces strict language identification: responses in the wrong language (e.g., Kyrgyz instead of Kazakh) receive Fluency = 0 regardless of content quality. For CLT conditions, only the final translated output is evaluated.

\section{Results}

\begin{table}[t]
\centering
\small
\caption{Aggregate performance across all eight models per condition. All scores are percentages of maximum possible. Time in seconds, length in words.}
\label{tab:conditions}
\begin{tabular}{@{}lcccccc@{}}
\toprule
\textbf{Condition} & \textbf{Total} & \textbf{Acc.} & \textbf{Flu.} & \textbf{Comp.} & \textbf{Time} & \textbf{Len.} \\
\midrule
C1: English          & 90.7 & 78.5 & 100.0 & 93.6 & 7.3  & 237 \\
\midrule
C2: Kazakh Dir.      & 76.9 & 65.5 & 89.0  & 76.1 & 14.6 & 128 \\
C3: Kazakh CLT       & 77.2 & 67.1 & 88.6  & 76.0 & 35.4 & 79  \\
\midrule
C4: Mongol.\ Dir.    & 74.0 & 62.7 & 87.5  & 71.8 & 25.0 & 143 \\
C5: Mongol.\ CLT     & 74.6 & 64.8 & 86.6  & 72.5 & 18.5 & 86  \\
\bottomrule
\end{tabular}
\end{table}

Table~\ref{tab:conditions} summarizes aggregate performance across all eight models for each condition. The English baseline (C1) achieves a mean score of 90.7\%, establishing an upper bound. Performance drops sharply for both low-resource languages: Kazakh direct (C2) scores 76.9\% and Mongolian direct (C4) scores 74.0\%---a gap of 13.8 and 16.7 percentage points respectively. Cross-lingual transfer yields modest gains: C3 (Kazakh CLT) outperforms C2 by +0.4pp, and C5 (Mongolian CLT) outperforms C4 by +0.6pp. Mean response time increases from 7.3s in English to 14.6s for Kazakh direct and 25.0s for Mongolian direct, while CLT conditions show mixed latency effects.

\begin{table}[t]
\centering
\small
\caption{Performance by model category. Eng.-First ($n$\!=\!250): Claude Opus 4.5, GPT-5.2, Gemini 3 Pro, Grok 4.1, Llama 4 Maverick. Bilingual ($n$\!=\!100): Qwen3, DeepSeek V3.2. Multiling.\ ($n$\!=\!50): Aya Expanse.}
\label{tab:categories}
\setlength{\tabcolsep}{4pt}
\begin{tabular}{@{}llrcrc@{}}
\toprule
\textbf{Cat.} & \textbf{Cond.} & \textbf{Score} & \textbf{$\Delta$CLT} & \textbf{Time} & \textbf{Len.} \\
\midrule
\multirow{5}{*}{\rotatebox[origin=c]{90}{\scriptsize\textbf{Eng.-First}}}
  & C1 & 92.2 & ---   & 7.2  & 231 \\
  & C2 & 88.2 & \multirow{2}{*}{$-$1.1} & 11.8 & 130 \\
  & C3 & 87.1 &       & 11.0 & 80  \\
  & C4 & 87.3 & \multirow{2}{*}{$-$0.7} & 25.4 & 137 \\
  & C5 & 86.7 &       & 14.7 & 89  \\
\midrule
\multirow{5}{*}{\rotatebox[origin=c]{90}{\scriptsize\textbf{Bilingual}}}
  & C1 & 91.2 & ---   & 6.4  & 251 \\
  & C2 & 79.2 & \multirow{2}{*}{+2.2} & 17.4 & 131 \\
  & C3 & 81.3 &       & 18.4 & 80  \\
  & C4 & 75.3 & \multirow{2}{*}{+4.3} & 29.1 & 163 \\
  & C5 & 79.7 &       & 30.5 & 90  \\
\midrule
\multirow{5}{*}{\rotatebox[origin=c]{90}{\scriptsize\textbf{Multiling.}}}
  & C1 & 82.3 & ---   & 10.1 & 239 \\
  & C2 & 15.7 & \multirow{2}{*}{+4.3} & 22.9 & 111 \\
  & C3 & 20.0 &       & 192\textsuperscript{\dag} & 71 \\
  & C4 &  4.7 & \multirow{2}{*}{$-$0.3} & 14.5 & 130 \\
  & C5 &  4.3 &       & 13.8 & 63  \\
\bottomrule
\multicolumn{6}{@{}l}{\textsuperscript{\dag}\footnotesize Inflated by pipeline timeout retries.}
\end{tabular}
\end{table}

Table~\ref{tab:categories} disaggregates results by model category. English-first models maintain strong performance across all conditions (86.7--92.2\%), with minimal variation between direct and CLT approaches. Bilingual models show the clearest CLT benefit: +2.2pp for Kazakh and +4.3pp for Mongolian. The multilingual model (Aya Expanse) scores 82.3\% on English but collapses to 15.7\% on Kazakh direct and 4.7\% on Mongolian direct, frequently producing Kyrgyz instead of Kazakh and garbled Mongolian output.

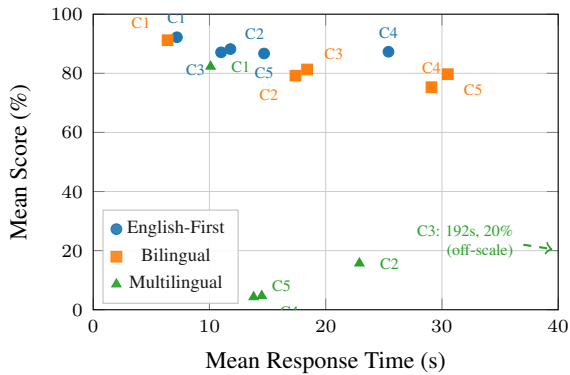
\begin{figure}[t]
\centering
\begin{tikzpicture}
\begin{axis}[
    width=\columnwidth,
    height=5.5cm,
    xlabel={Mean Response Time (s)},
    ylabel={Mean Score (\%)},
    xmin=0, xmax=40,
    ymin=0, ymax=100,
    grid=both,
    grid style={line width=0.1pt, draw=gray!20},
    major grid style={line width=0.2pt, draw=gray!40},
    legend style={
        at={(0.02,0.02)},
        anchor=south west,
        font=\scriptsize,
        draw=gray!50,
        fill=white,
        fill opacity=0.9,
    },
    tick label style={font=\scriptsize},
    label style={font=\small},
    every axis plot/.append style={only marks, mark size=2pt},
]
\addplot[color=engfirst, mark=*, mark options={fill=engfirst}]
    coordinates {(7.2,92.2)(11.8,88.2)(11.0,87.1)(25.4,87.3)(14.7,86.7)};
\addlegendentry{English-First}
\addplot[color=bilingual, mark=square*, mark options={fill=bilingual}]
    coordinates {(6.4,91.2)(17.4,79.2)(18.4,81.3)(29.1,75.3)(30.5,79.7)};
\addlegendentry{Bilingual}
\addplot[color=multilingual, mark=triangle*, mark options={fill=multilingual}]
    coordinates {(10.1,82.3)(22.9,15.7)(14.5,4.7)(13.8,4.3)};
\addlegendentry{Multilingual}
\node[font=\tiny,color=engfirst,anchor=south] at (axis cs:7.2,93.5) {C1};
\node[font=\tiny,color=engfirst,anchor=south west] at (axis cs:12.2,88.2) {C2};
\node[font=\tiny,color=engfirst,anchor=north east] at (axis cs:10.5,86.5) {C3};
\node[font=\tiny,color=engfirst,anchor=south] at (axis cs:25.4,88.5) {C4};
\node[font=\tiny,color=engfirst,anchor=north] at (axis cs:14.7,85.5) {C5};
\node[font=\tiny,color=bilingual,anchor=south east] at (axis cs:5.8,91.8) {C1};
\node[font=\tiny,color=bilingual,anchor=north east] at (axis cs:16.8,78.2) {C2};
\node[font=\tiny,color=bilingual,anchor=south west] at (axis cs:19.0,81.3) {C3};
\node[font=\tiny,color=bilingual,anchor=south] at (axis cs:29.1,76.5) {C4};
\node[font=\tiny,color=bilingual,anchor=north west] at (axis cs:31.0,79.7) {C5};
\node[font=\tiny,color=multilingual,anchor=west] at (axis cs:11.0,82.3) {C1};
\node[font=\tiny,color=multilingual,anchor=west] at (axis cs:23.8,15.7) {C2};
\node[font=\tiny,color=multilingual,anchor=north west] at (axis cs:15.2,4.7) {C4};
\node[font=\tiny,color=multilingual,anchor=south west] at (axis cs:14.5,3.0) {C5};
\draw[->,thick,color=multilingual,dashed] (axis cs:37,22) -- (axis cs:39.5,20.5);
\node[font=\tiny,color=multilingual,anchor=east,align=right] at (axis cs:37,23) {C3: 192s, 20\%\\(off-scale)};
\end{axis}
\end{tikzpicture}
\caption{Mean score vs.\ mean response time by model category and condition. The Multilingual C3 point (191.9s, 20.0\%) is off-scale and annotated.}
\label{fig:scatter}
\end{figure}

Figure~\ref{fig:scatter} plots mean score against mean response time for each condition--category pair. English-first models cluster in the high-accuracy, low-latency region across all conditions. Bilingual models occupy a middle band with notably higher latency for Mongolian. Aya Expanse appears as a clear outlier in all non-English conditions, combining the lowest scores with variable response times.

\section{Discussion}

\paragraph{The Fluency Illusion.}
A striking pattern emerges from the sub-metric breakdown: Fluency degrades far less than Accuracy when moving from English to low-resource languages. For Kazakh, Accuracy drops 13.0 percentage points (from 78.5\% to 65.5\%) while Fluency drops only 11.0pp (from 100\% to 89.0\%). The gap widens for Mongolian---Accuracy falls 16.0pp to 62.5\% while Fluency remains at 87.5\%. Models generate text that reads naturally in the target language while containing factual errors or hallucinations---a potentially deceptive failure mode for end users who lack access to English-language verification.

\paragraph{CLT as Selective Scaffolding.}
Cross-lingual transfer does not uniformly improve performance. English-first models, which likely already leverage internal English representations, show negligible or slightly negative effects from the explicit CLT pipeline ($-$1.1pp for Kazakh, $-$0.7pp for Mongolian). By contrast, bilingual models benefit meaningfully (+2.2pp for Kazakh, +4.3pp for Mongolian), suggesting that the structured decomposition---translate, reason, translate back---compensates for weaker reasoning capabilities in the target language. This positions CLT not as a universal strategy but as targeted scaffolding for models with partial multilingual competence.

\paragraph{Multilingual in Name Only.}
Aya Expanse, explicitly designed for 100+ languages including Turkic family members, achieves 82.3\% in English but collapses to 15.7\% for Kazakh and 4.7\% for Mongolian. The significant drop is mostly due to the model frequently producing Kyrgyz---a closely related but distinct language---instead of Kazakh, evidenced by the systematic absence of Kazakh-specific characters and the regular use of Kyrgyz words. This suggests that broad language coverage in training does not guarantee reliable production of individual low-resource languages, particularly within closely related language families.

Our study is limited by its benchmark size (50 questions) and reliance on semi-automated grading with a single-person team, though the use of a validated answer sheet with strict rubric enforcement mitigates the latter concern.

\section{Conclusion}

Our benchmark of 2,000 LLM responses across Kazakh and Mongolian reveals a consistent and substantial performance gap between English and low-resource language conditions, with models scoring 13.8--16.7 percentage points lower when prompted in LRLs. Cross-lingual transfer provides measurable but selective gains---benefiting bilingual architectures while offering little to models that already reason through English internally. Perhaps most notably, the only model explicitly designed for multilingual coverage performs worst on the languages it claims to support, producing a related but wrong language instead.

These findings underscore an urgent need for language-specific evaluation beyond aggregate multilingual benchmarks. Future work should expand this framework to additional Turkic and Mongolic languages, and investigate whether targeted fine-tuning on small LRL corpora can close the gap more effectively than prompting strategies alone.

\bibliography{references}

\end{document}